\documentclass[11pt,a4paper]{article}

\PassOptionsToPackage{hyphens}{url}

\usepackage{naaclhlt2018}

\usepackage{times}
\usepackage{latexsym}
\usepackage{graphicx}
\usepackage{url}
\usepackage[utf8]{inputenc}

\aclfinalcopy 

\title{A Portuguese Native Language Identification Dataset}

\author{Iria del Río\textsuperscript{1}, Marcos Zampieri\textsuperscript{2}, Shervin Malmasi\textsuperscript{3,4} \\
\textsuperscript{1}University of Lisbon, Center of Linguistics-CLUL, Portugal\\
\textsuperscript{2}University of Wolverhampton, United Kingdom\\
\textsuperscript{3}Harvard Medical School, United States\\
\textsuperscript{4}Macquarie University, Australia\\
  {\tt igayo@letras.ulisboa.pt}
}

\date{}

\begin{document}
\maketitle
\begin{abstract}
In this paper we present NLI-PT, the first Portuguese dataset compiled for Native Language Identification (NLI), the task of identifying an author's first language based on their second language writing. The dataset includes 1,868 student essays written by learners of European Portuguese, native speakers of the following L1s: Chinese, English, Spanish, German, Russian, French, Japanese, Italian, Dutch, Tetum, Arabic, Polish, Korean, Romanian, and Swedish. NLI-PT includes the original student text and four different types of annotation: POS, fine-grained POS, constituency parses, and dependency parses. NLI-PT can be used not only in NLI but also in research on several topics in the field of Second Language Acquisition and educational NLP. We discuss possible applications of this dataset and present the results obtained for the first lexical baseline system for Portuguese NLI.
\end{abstract}

\section{Introduction}

Several learner corpora have been compiled for English, such as the International Corpus of Learner English \cite{gran2003}. The importance of such resources has been increasingly recognized across a variety of research areas, from Second Language Acquisition to Natural Language Processing. Recently, we have seen substantial growth in this area and new corpora for languages other than English have appeared. For Romance languages, there are a several corpora and resources for French\footnote{https://uclouvain.be/en/research-institutes/ilc/cecl/frida.html}, Spanish \cite{lozano2010}, and Italian \cite{boyd2014}. 

Portuguese has also received attention in the compilation of learner corpora. There are two corpora compiled at the School of Arts and Humanities of the University of Lisbon: the corpus \textit{Recolha de dados de Aprendizagem do Portugu{\^e}s L{\'{i}}ngua Estrangeira}\footnote{http://www.clul.ulisboa.pt/pt/24-recursos/350-recolha-de-dados-de-ple} (hereafter, Leiria corpus), with 470 texts and 70,500 tokens, and the Learner Corpus of Portuguese as Second/Foreign Language, COPLE2\footnote{http://alfclul.clul.ul.pt/teitok/learnercorpus} \cite{rio2016}, with 1,058 texts and 201,921 tokens. The \textit{Corpus de Produ{\c c}{\~o}es Escritas de Aprendentes de PL2}, PEAPL2\footnote{http://teitok.iltec.pt/peapl2/} compiled at the University of Coimbra, contains 516 texts and 119,381 tokens. Finally, the \textit{Corpus de Aquisi{\c c}{\~a}o de L2}, CAL2\footnote{http://cal2.clunl.edu.pt/}, compiled at the New University of Lisbon, contains 1,380 texts and 281,301 words, and it includes texts produced by adults and children, as well as a spoken subset. 

The aforementioned Portuguese learner corpora contain very useful data for research,   particularly for Native Language Identification (NLI), a task that has received much attention in recent years.
NLI is the task of determining the native language (L1) of an author based on their second language (L2) linguistic productions \cite{malmasi:2017:nlisg}. NLI works by identifying language use patterns that are common to groups of speakers of the same native language. This process is underpinned by the presupposition that an author’s L1 disposes them towards certain language production patterns in their L2, as influenced by their mother tongue.
A major motivation for NLI is studying second language acquisition. NLI models can enable analysis of inter-L1 linguistic differences, allowing us to study the language learning process and develop L1-specific pedagogical methods and materials.

However, there are limitations to using existing Portuguese data for NLI.
An important issue is that the different corpora each contain data collected from different L1 backgrounds in varying amounts; they would need to be combined to have sufficient data for an NLI study.
Another challenge concerns the annotations as only two of the corpora (PEAPL2 and COPLE2) are linguistically annotated, and this is limited to POS tags. 
The different data formats used by each corpus presents yet another challenge to their usage.

In this paper we present NLI-PT, a dataset collected for Portuguese NLI. The dataset is made freely available for research purposes.\footnote{NLI-PT is available at: http://www.clul.ulisboa.pt/en/resources-en/11-resources/894-nli-pt-a-portuguese-native-language-identification-dataset}
With the goal of unifying learner data collected from various sources, listed in  Section \ref{sec:methodology}, we applied a methodology which has been previously used for the compilation of language variety corpora \cite{tan2014}.
The data was converted to a unified data format and uniformly annotated at different linguistic levels as described in Section \ref{sec:preprocessing}.
To the best of our knowledge, NLI-PT is the only Portuguese dataset developed specifically for NLI, this will open avenues for research in this area.

\section{Related Work}

NLI has attracted a lot of attention in recent years. Due to the availability of suitable data, as discussed earlier, this attention has been particularly focused on English. The most notable examples are the two editions of the NLI shared task organized in 2013 \cite{tetreault-blanchard-cahill:2013:BEA} and 2017 \cite{nli2017}.

Even though most NLI research has been carried out on English data, an important research trend in recent years has been the application of NLI methods to other languages, as discussed in \newcite{multilingual-nli}. Recent NLI studies on languages other than English include Arabic \cite{malmasi:2014:anli} and Chinese \cite{malmasi:2014:cnli,wang-malmasi-huang:2015}. To the best of our knowledge, no study has been published on Portuguese and the NLI-PT dataset opens new possibilities of research for Portuguese. In Section \ref{sec:NLI} we present the first simple baseline results for this task.

Finally, as NLI-PT can be used in other applications besides NLI, it is important to point out that a number of studies have been published on educational NLP applications for Portuguese and on the compilation of learner language resources for Portuguese. Examples of such studies include grammatical error correction \cite{martins1998linguistic}, automated essay scoring \cite{elliot2003intellimetric}, academic word lists \cite{baptista2010p}, and the learner corpora presented in the previous section.

\section{Corpus Description}
\subsection{Collection methodology}
\label{sec:methodology}

The data was collected from three different learner corpora of Portuguese: (i) COPLE2; (ii) Leiria corpus, and (iii) PEAPL2\footnote{In the near future we want to incorporate also data from the CAL2 corpus.} as presented in Table~\ref{tab:data}.

\setlength{\tabcolsep}{2pt}
\def\arraystretch{1}

\begin{table}[!ht]
\begin{center}
\scalebox{0.95}{
\begin{tabular}{lcccc}
\hline\noalign{\smallskip}
\bf  & \bf COPLE2 & \bf LEIRIA & \bf PEAPL2 &\bf TOTAL \\
\noalign{\smallskip}
\hline
Texts & 1,058 & 330 & 480 & 1,868 \\
Tokens & 201,921 & 57,358 & 121,138 & 380,417 \\
Types & 9,373 & 4,504 & 6,808 & 20,685 \\
TTR & 0.05 & 0.08 & 0.06 & 0.05 \\
\hline
\end{tabular}
}
\vspace{3mm}
\caption{Distribution of the dataset: Number of texts, tokens, types, and type/token ratio (TTER) per source corpus.}
\label{tab:data}
\end{center}
\end{table}

\noindent The three corpora contain written productions from learners of Portuguese with different proficiency levels and native languages (L1s). In the dataset we included all the data in COPLE2 and sections of PEAPL2 and Leiria corpus. 

\begin{figure*}[!ht]
\centering
\includegraphics[width=.68\textwidth]{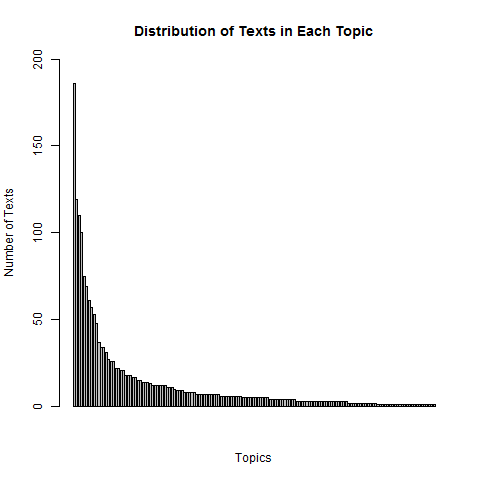}
\vspace{-2mm}
\caption{Topic distribution by number of texts. Each bar represents one of the 148 topics.}
\label{fig:topics}
\end{figure*}

The main variable we used for text selection was the presence of specific L1s. Since the three corpora consider different L1s, we decided to use the L1s present in the largest corpus, COPLE2, as the reference. Therefore, we included in the dataset texts corresponding to the following 15 L1s: Chinese, English, Spanish, German, Russian, French, Japanese, Italian, Dutch, Tetum, Arabic, Polish, Korean, Romanian, and Swedish. It was the case that some of the L1s present in COPLE2 were not documented in the other corpora. The number of texts from each L1 is presented in Table~\ref{tab:l1}.

\setlength{\tabcolsep}{2pt}
\def\arraystretch{1}

\begin{table}[!ht]
\begin{center}
\scalebox{0.93}{
\begin{tabular}{lcccc}
\hline\noalign{\smallskip}
\bf  & \bf COPLE2 & \bf PEAPL2 & \bf LEIRIA  &\bf TOTAL \\
\noalign{\smallskip}
\hline
\noalign{\smallskip}
Arabic & 13 & 1 & 0 & 14 \\
Chinese & 323 & 32 & 0 & 355 \\
Dutch & 17 & 26 & 0 & 43 \\
English & 142 & 62 & 31 & 235 \\
French & 59 & 38 & 7 & 104 \\
German	& 86 & 88 & 40 & 214 \\
Italian & 49 & 83 & 83 & 215 \\
Japanese & 52 & 15 & 0 & 67 \\
Korean	& 9 & 9 & 48 & 66 \\
Polish 	& 31 & 28 & 12 & 71 \\
Romanian & 12 & 16 & 51 & 79 \\
Russian & 80 & 11 & 1 & 92 \\
Spanish & 147 & 68 & 56 & 271 \\
Swedish	& 16 & 2 & 1 & 19 \\
Tetum & 22 & 1 & 0 & 23 \\
\hline
 Total & 1,058 & 480 & 330 & 1,868 \\
\hline
\end{tabular}
}
\vspace{3mm}
\caption{Distribution by L1s and source corpora.}
\label{tab:l1}
\end{center}
\end{table}

Concerning the corpus design, there is some variability among the sources we used. Leiria corpus and PEAPL2 followed a similar approach for data collection and show a close design. They consider a close list of topics, called “stimulus”, which belong to three general areas: (i) the individual; (ii) the society; (iii) the environment. Those topics are presented to the students in order to produce a written text.  As a whole, texts from PEAPL2 and Leiria represent 36 different stimuli or topics in the dataset. In COPLE2 corpus the written texts correspond to written exercises done during Portuguese lessons, or to official Portuguese proficiency tests. For this reason, the topics considered in COPLE2 corpus are different from the topics in Leiria and PEAPL2. The number of topics is also larger in COPLE2 corpus: 149 different topics. There is some overlap between the different topics considered in COPLE2, that is, some topics deal with the same subject. This overlap allowed us to reorganize COPLE2 topics in our dataset, reducing them to 112.

\begin{figure*}[!ht]
\centering
\includegraphics[width=.65\textwidth]{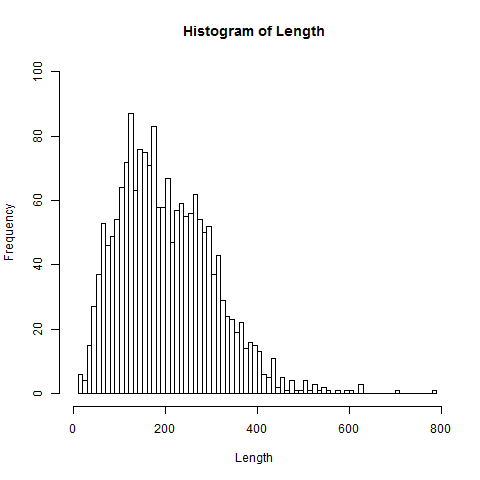}
\vspace{-2mm}
\caption{Histogram of document lengths, as measured by the number of tokens. The mean value is 204 with standard deviation of 103.}
\label{fig:histogram}
\end{figure*}

\setlength{\tabcolsep}{2pt}
\def\arraystretch{1}

\begin{table}[!ht]
\begin{center}
\scalebox{0.95}{
\begin{tabular}{lc}
\hline\noalign{\smallskip}
\bf  & \bf Number of topics \\
\noalign{\smallskip}
\hline
COPLE2 & 112 \\
PEAPL2+Leiria & 36 \\
Total & 148 \\
\hline
\end{tabular}
}
\vspace{3mm}
\caption{Number of different topics by source.}
\label{tab:data}
\end{center}
\end{table}

\noindent Due to the different distribution of topics in the source corpora, the 148 topics in the dataset are not represented uniformly. Three topics account for a 48.7\% of the total texts and, on the other hand, a 72\% of the topics are represented by 1-10 texts (Figure \ref{fig:topics}). This variability affects also text length. The longest text has 787 tokens and the shortest has only 16 tokens. Most texts, however, range roughly from 150 to 250 tokens. To better understand the distribution of texts in terms of word length we plot a histogram of all texts with their word length in bins of 10 (1-10 tokens, 11-20 tokens, 21-30 tokens and so on) (Figure \ref{fig:histogram}).

The three corpora use the proficiency levels defined in the Common European Framework of Reference for Languages (CEFR), but they show differences in the number of levels they consider. There are five proficiency levels in COPLE2 and PEAPL2: A1, A2, B1, B2, and C1. But there are 3 levels in Leiria corpus: A, B, and C. The number of texts included from each proficiency level is presented in Table \ref{tab:proficiency}.

\setlength{\tabcolsep}{2pt}
\def\arraystretch{1}

\begin{table}[!ht]
\centering
\scalebox{0.95}{
\begin{tabular}{lcccc}
\hline\noalign{\smallskip}
\bf  & \bf COPLE2 & \bf LEIRIA & \bf PEAPL2 & \bf TOTAL \\
\noalign{\smallskip}
\hline
A1 & 91 & n/a & 78 & 169 \\
A2 & 414 & n/a &  89 & 503 \\
A & 505 & 203 & 167 & 875 \\
\hline
B1 & 312 & n/a &  203 & 515 \\
B2 & 202 & n/a &  70 & 272 \\
B & 514 & 89 & 273 & 876 \\
\hline
C1 & 39 & n/a &  40 & 79 \\
C & 39 & 38 & 40 & 117 \\
\hline
\end{tabular}
}
\vspace{2mm}
\caption{Distribution by proficiency levels and by source corpus.}
\label{tab:proficiency}
\end{table}

\subsection{Preprocessing and annotation of texts}
\label{sec:preprocessing}

As demonstrated earlier, these learner corpora use different formats. COPLE2 is mainly codified in XML, although it gives the possibility of getting the student version of the essay in TXT format. PEAPL2 and Leiria corpus are compiled in TXT format.\footnote{Currently there is a XML version of PEAPL2, but this version was not available when we compiled the dataset.} In both corpora, the TXT files contain the student version with special annotations from the transcription.
For the NLI experiments we were interested in a clean txt version of the students' text, together with versions annotated at different linguistics levels. Therefore, as a first step, we removed all the annotations corresponding to the transcription process in PEAPL2 and Leiria files. As a second step, we proceeded to the linguistic annotation of the texts using different NLP tools. 

We annotated the dataset at two levels: Part of Speech (POS) and syntax. We performed the annotation with freely available tools for the Portuguese language. For POS we added a simple POS, that is, only type of word, and a fine-grained POS, which is the type of word plus its morphological features. We used the LX Parser \cite{silva2010}, for the simple POS and the Portuguese morphological module of Freeling \mbox{\cite{padro2012}}, for detailed POS. Concerning syntactic annotations, we included constituency and dependency annotations. For constituency parsing, we used the LX Parser, and for dependency, the DepPattern toolkit \cite{gamallo2012}.

\section{Applications}
\label{sec:applications}

NLI-PT was developed primarily for NLI, but it can be used for other research purposes ranging from second language acquisition to educational NLP applications. Here are a few examples of applications in which the dataset can be used:

\begin{itemize}
\item Computer-aided Language Learning (CALL): CALL software has been developed for Portuguese \cite{marujo2009porting}. Further improvements in these tools can take advantage of the training material available in NLI-PT for a number of purposes such as L1-tailored exercise design. 

\item Grammatical error detection and correction: as discussed in \newcite{zampieri2014grammatical}, a known challenge in this task is acquiring suitable training data to account for the variation of errors present in non-native texts. One of the strategies developed to cope with this problem is to generate artificial training data \cite{felice2014generating}. Augmenting training data using a suitable annotated dataset such as NLI-PT can improve the quality of existing grammatical error correction systems for Portuguese.

\item Spellchecking: Studies have shown that general-purpose spell checkers target performance errors but fail to address many competence errors committed by language learners \cite{rimrott2005language}. To address this shortcoming a number of spell checking tools have been developed for language learners \cite{ndiaye2003spell}. Suitable training data is required o develop these tools. NLI-PT is a suitable resource to train learner spell checkers for Portuguese.

\item L1 interference: one of the aspects of non-native language production that can be studied using data-driven methods is the influence of L1 in non-native speakers production. Its annotation and the number of second languages included in the dataset make NLI-PT a perfect fit for such studies.
\end{itemize}

\subsection{A Baseline for Portuguese NLI}
\label{sec:NLI}

To demonstrate the usefulness of the dataset we present the first lexical baseline for Portuguese NLI using a sub-set of NLI-PT. To the best of our knowledge, no study has been published on Portuguese NLI and our work fills this gap. 

In this experiment we included the five L1s in NLI-PT which contain the largest number of texts in this sub-set and run a simple linear SVM \cite{liblinear} classifier using a bag of words model to identify the L1 of each text. The languages included in this experiment were Chinese (355 texts), English (236 texts), German (214 texts), Italian (216 texts), and Spanish (271 texts). 

We evaluated the model using stratified 10-fold cross-validation, achieving 70\% accuracy. 
An important limitation of this experiment is that it does not account for topic bias, an important issue in NLI \cite{malmasi2016}. This is due to the fact that NLI-PT is not balanced by topic and the model could be learning topic associations instead.\footnote{See \newcite[p. 23]{malmasi2016} for a detailed discussion.}
In future work we would like to carry out using syntactic features such as function words, syntactic relations and POS annotation.

\section{Conclusion and Future Work}

This paper presented NLI-PT, the first Portuguese dataset compiled for NLI. NLI-PT contains 1,868 texts written by speakers of 15 L1s amounting to over 380,000 tokens. 

As discussed in Section \ref{sec:applications}, NLI-PT opens several avenues for future research. It can be used for different research purposes beyond NLI such as grammatical error correction and CALL. An experiment with the texts written by the speakers of five L1s: Chinese, English, German, Italian, and Spanish using a bag of words model achieved 70\% accuracy. We are currently experimenting with different features taking advantage of the annotation available in NLI-PT thus reducing topic bias in classification.

In future work we would like to include more texts in the dataset following the same methodology and annotation.

\section*{Acknowledgement}
We want to thank the research teams that have made available the data we used in this work: Centro de Estudos de Linguística Geral e Aplicada at Universidade de Coimbra (specially Cristina Martins) and Centro de Linguística da Universidade de Lisboa (particularly Amália Mendes).

This work was partially supported by Fundação para a Ciência e a Tecnologia (postdoctoral research grant SFRH/BPD/109914/2015).


\bibliographystyle{acl_natbib}
\bibliography{naaclhlt2018,nlipt}

\begin{thebibliography}{25}
\expandafter\ifx\csname natexlab\endcsname\relax\def\natexlab#1{#1}\fi

\bibitem[{Baptista et~al.(2010)Baptista, Costa, Guerra, Zampieri, Cabral, and
  Mamede}]{baptista2010p}
Jorge Baptista, Neuza Costa, Joaquim Guerra, Marcos Zampieri, Maria Cabral, and
  Nuno Mamede. 2010.
\newblock {P-AWL: academic word list for Portuguese}.
\newblock In \emph{Proceedings of PROPOR}.

\bibitem[{Boyd et~al.(2014)Boyd, Hana, Nicolas, Meurers, Wisniewski, andrea
  Abel, Schöne, Štindlová, and Vettori}]{boyd2014}
Adriane Boyd, Jirka Hana, Lionel Nicolas, Detmar Meurers, Katrin Wisniewski,
  andrea Abel, Karin Schöne, Barbora Štindlová, and Chiara Vettori. 2014.
\newblock {The MERLIN corpus: Learner Language and the CEFR}.
\newblock In \emph{Proceedings of LREC}.

\bibitem[{Elliot(2003)}]{elliot2003intellimetric}
Scott Elliot. 2003.
\newblock {IntelliMetric: From here to validity}.
\newblock \emph{Automated essay scoring: A cross-disciplinary perspective},
  pages 71--86.

\bibitem[{Fan et~al.(2008)Fan, Chang, Hsieh, Wang, and Lin}]{liblinear}
Rong-En Fan, Kai-Wei Chang, Cho-Jui Hsieh, Xiang-Rui Wang, and Chih-Jen Lin.
  2008.
\newblock {LIBLINEAR: A Library for Large Linear Classification}.
\newblock \emph{Journal of Machine Learning Research}, 9(Aug):1871--1874.

\bibitem[{Felice and Yuan(2014)}]{felice2014generating}
Mariano Felice and Zheng Yuan. 2014.
\newblock {Generating Artificial Errors for Grammatical Error Correction}.
\newblock In \emph{Proceedings of the EACL Student Research Workshop}.

\bibitem[{Granger(2003)}]{gran2003}
Sylviane Granger. 2003.
\newblock The international corpus of learner english: A new resource for
  foreign language learning and teaching and second language acquisition
  research.
\newblock \emph{TESOL Quarterly}, 37(3):538--546.

\bibitem[{Lozano(2010)}]{lozano2010}
Crist{\'o}bal Lozano. 2010.
\newblock \emph{{CEDEL}2, {Corpus} {Escrito} del {Espa{\~n}ol} {L}2}.
\newblock Department of English, Universidad Aut{\'o}noma de Madrid, Madrid.

\bibitem[{Malmasi(2016)}]{malmasi2016}
Shervin Malmasi. 2016.
\newblock \emph{{Native Language Identification: Explorations and
  Applications}}.
\newblock Ph.D. thesis.

\bibitem[{Malmasi and Dras(2014{\natexlab{a}})}]{malmasi:2014:anli}
Shervin Malmasi and Mark Dras. 2014{\natexlab{a}}.
\newblock {Arabic Native Language Identification}.
\newblock In \emph{Proceedings of the Arabic Natural Language Processing
  Workshop}.

\bibitem[{Malmasi and Dras(2014{\natexlab{b}})}]{malmasi:2014:cnli}
Shervin Malmasi and Mark Dras. 2014{\natexlab{b}}.
\newblock {Chinese Native Language Identification}.
\newblock In \emph{Proceedings of EACL}.

\bibitem[{Malmasi and Dras(2015)}]{multilingual-nli}
Shervin Malmasi and Mark Dras. 2015.
\newblock {Multilingual Native Language Identification}.
\newblock In \emph{Natural Language Engineering}.

\bibitem[{Malmasi and Dras(2017)}]{malmasi:2017:nlisg}
Shervin Malmasi and Mark Dras. 2017.
\newblock {Native Language Identification using Stacked Generalization}.
\newblock \emph{arXiv preprint arXiv:1703.06541}.

\bibitem[{Malmasi et~al.(2017)Malmasi, Evanini, Cahill, Tetreault, Pugh,
  Hamill, Napolitano, and Qian}]{nli2017}
Shervin Malmasi, Keelan Evanini, Aoife Cahill, Joel Tetreault, Robert Pugh,
  Christopher Hamill, Diane Napolitano, and Yao Qian. 2017.
\newblock {A Report on the 2017 Native Language Identification Shared Task}.
\newblock In \emph{Proceedings of BEA}.

\bibitem[{Martins et~al.(1998)Martins, Hasegawa, Nunes, Montilha, and
  De~Oliveira}]{martins1998linguistic}
Ronaldo~Teixeira Martins, Ricardo Hasegawa, Maria das Gra{\c{c}}as~Volpe Nunes,
  Gisele Montilha, and Osvaldo~Novais De~Oliveira. 1998.
\newblock {Linguistic issues in the development of ReGra: A grammar checker for
  Brazilian Portuguese}.
\newblock \emph{Natural Language Engineering}, 4(4):287--307.

\bibitem[{Marujo et~al.(2009)Marujo, Lopes, Mamede, Trancoso, Pino, Eskenazi,
  Baptista, and Viana}]{marujo2009porting}
Lu{\'\i}s Marujo, Jos{\'e} Lopes, Nuno Mamede, Isabel Trancoso, Juan Pino,
  Maxine Eskenazi, Jorge Baptista, and C{\'e}u Viana. 2009.
\newblock {Porting REAP to European Portuguese}.
\newblock In \emph{Proceedings of the International Workshop on Speech and
  Language Technology in Education}.

\bibitem[{Ndiaye and Faltin(2003)}]{ndiaye2003spell}
Mar Ndiaye and Anne~Vandeventer Faltin. 2003.
\newblock {A Spell Checker Tailored to Language Learners}.
\newblock \emph{Computer Assisted Language Learning}, 16(2-3):213--232.

\bibitem[{Otero and Gonz{\'{a}}lez(2012)}]{gamallo2012}
Pablo~Gamallo Otero and Isaac Gonz{\'{a}}lez. 2012.
\newblock {DepPattern: a Multilingual Dependency Parser}.
\newblock In \emph{Proceedings of PROPOR}.

\bibitem[{Padr{\'{o}} and Stanilovsky(2012)}]{padro2012}
Llu{\'{i}}s Padr{\'{o}} and Evgeny Stanilovsky. 2012.
\newblock Freeling 3.0: Towards wider multilinguality.
\newblock In \emph{Proceedings LREC}.

\bibitem[{Rimrott and Heift(2005)}]{rimrott2005language}
Anne Rimrott and Trude Heift. 2005.
\newblock {Language Learners and Generic Spell Checkers in CALL}.
\newblock \emph{CALICO journal}, pages 17--48.

\bibitem[{del Río et~al.(2016)del Río, Antunes, Mendes, and
  Janssen}]{rio2016}
Iria del Río, Sandra Antunes, Amália Mendes, and Maarten Janssen. 2016.
\newblock Towards error annotation in a learner corpus of portuguese.
\newblock In \emph{Proceedings of the NLP4CALL workshop at SLTC}, pages 8--17.

\bibitem[{Silva et~al.(2010)Silva, Branco, Castro, and Reis}]{silva2010}
Jo{\~{a}}o~Ricardo Silva, Ant{\'{o}}nio Branco, S{\'{e}}rgio Castro, and Ruben
  Reis. 2010.
\newblock {Out-of-the-Box Robust Parsing of Portuguese}.
\newblock In \emph{Proceedings of PROPOR}, pages 75--85.

\bibitem[{Tan et~al.(2014)Tan, Zampieri, Ljube{\v{s}}ic, and
  Tiedemann}]{tan2014}
Liling Tan, Marcos Zampieri, Nikola Ljube{\v{s}}ic, and J{\"o}rg Tiedemann.
  2014.
\newblock {Merging Comparable Data Sources for the Discrimination of Similar
  Languages: The DSL Corpus Collection}.
\newblock In \emph{Proceedings of the BUCC Workshop}.

\bibitem[{Tetreault et~al.(2013)Tetreault, Blanchard, and
  Cahill}]{tetreault-blanchard-cahill:2013:BEA}
Joel Tetreault, Daniel Blanchard, and Aoife Cahill. 2013.
\newblock A report on the first native language identification shared task.
\newblock In \emph{Proceedings of BEA}.

\bibitem[{Wang et~al.(2015)Wang, Malmasi, and Huang}]{wang-malmasi-huang:2015}
Maolin Wang, Shervin Malmasi, and Mingxuan Huang. 2015.
\newblock {The Jinan Chinese Learner Corpus}.
\newblock In \emph{Proceedings of BEA}.

\bibitem[{Zampieri and Tan(2014)}]{zampieri2014grammatical}
Marcos Zampieri and Liling Tan. 2014.
\newblock {Grammatical Error Detection with Limited Training Data: The Case of
  Chinese}.
\newblock In \emph{Proceedings of ICCE}.

\end{thebibliography}

\end{document}